\documentclass{sigchi}

\CopyrightYear{2020}

\usepackage{balance}       
\usepackage{graphics}      
\usepackage[T1]{fontenc}   
\usepackage{txfonts}
\usepackage{mathptmx}
\usepackage[pdflang={en-US},pdftex]{hyperref}
\usepackage{color}
\usepackage{booktabs}
\usepackage{textcomp}

\usepackage{amsmath}
\usepackage{booktabs}
\usepackage{algorithm}
\usepackage{algorithmic}
\usepackage{epsfig}
\usepackage{amssymb}
\usepackage{caption}
\usepackage[shortlabels]{enumitem}
\usepackage{microtype}        
\usepackage{ccicons}          

\usepackage{todonotes}

\def\plaintitle{ImagineNet: Restyling Apps Using Neural Style Transfer}
\def\plainauthor{Michael H. Fischer, Richard R. Yang, Monica S. Lam}

\def\plainkeywords{Style transfer; graphical user interfaces; automated design.}

\makeatletter
\def\url@leostyle{%
  \@ifundefined{selectfont}{
    \def\UrlFont{\sf}
  }{
    \def\UrlFont{\small\bf\ttfamily}
  }}
\makeatother
\def\pprw{8.5in}
\def\pprh{11in}

\definecolor{linkColor}{RGB}{6,125,233}
\hypersetup{%
  pdftitle={\plaintitle},
 pdfauthor={\plainauthor},
  pdfkeywords={\plainkeywords},
  pdfdisplaydoctitle=true, 
  bookmarksnumbered,
  pdfstartview={FitH},
  colorlinks,
  citecolor=black,
  filecolor=black,
  linkcolor=black,
  urlcolor=linkColor,
  breaklinks=true,
  hypertexnames=false
}

\begin{document}
\bibliographystyle{SIGCHI-Reference-Format}

\title{\plaintitle}

\numberofauthors{1}
\author{%
  \alignauthor{Michael H. Fischer, Richard R. Yang, Monica S. Lam\\
    \affaddr{Computer Science Department}\\
    \affaddr{Stanford University}\\
    \affaddr{Stanford, CA 94309}\\
    \email{\{mfischer, richard.yang, lam\}@cs.stanford.edu}\\
    }
  }


\maketitle
\begin{abstract}

This paper presents ImagineNet, a tool that uses a novel neural style transfer model to enable end-users and app developers to restyle GUIs using an image of their choice. Former neural style transfer techniques are inadequate for this application because they produce GUIs that are illegible and hence nonfunctional. We propose a neural solution by adding a new loss term to the original formulation, which minimizes the squared error in the uncentered cross-covariance of features from different levels in a CNN between the style and output images. ImagineNet retains the details of GUIs, while transferring the colors and textures of the art. 
We presented GUIs restyled with ImagineNet as well as other style transfer techniques to 50 evaluators and all preferred those of ImagineNet. 
We show how ImagineNet can be used to restyle (1) the graphical assets of an app, (2) an app with user-supplied content, and (3) an app with dynamically generated GUIs.




\end{abstract}




\newcommand{\Lcontent}[0]{L_\textsc{content}}
\newcommand{\Ltexture}[0]{L_\textsc{texture}}
\newcommand{\Lstructure}[0]{L_\textsc{structure}}
\newcommand{\LTV}[0]{L_\textsc{TV}}
\newcommand{\TODO}[1]{{\color{red} TODO: #1}}
\newcommand{\short}[1]{{}}

\section{Introduction}
While the majority of mobile apps have ``one-size-fits-all'' graphical user interfaces (GUIs), many users enjoy variety in their GUIs \cite{weld2003automatically, norton2012ikea}.
Variety can be introduced by the app maker. For example, Candy Crush uses different styles at different stages of their game, or Line messenger allows changing the start screen, friends list, and chat screens.
Variety can also be controlled by the users.
Application launchers for Android allow users to choose from a wide collection of third-party themes that change common apps and icons to match the theme of choice.
These stylizing approaches require hardcoding specific elements for a programming framework; they don't generalize and are brittle if the developers change their application.

Our goal is to develop a system that enables content creators or end users to restyle an app interface with the artwork of their choosing, effectively bringing the skill of renowned artists to their fingertips.
We imagine a future where users will expect to see beautiful designs in every app they use, and to enjoy variety in design like they would with fashion today.

Gatys et al.~\cite{1508.06576} proposed neural style transfer as a method to turn images into art of a given style.
Leveraging human creativity and artistic ability, style transfer technique generates results that are richer, more beautiful and varied than the predictable outcomes produced with traditional image transforms such as blurring, sharpening, and color transforms.  
The insight behind the original style transfer algorithm is to leverage the pre-trained representation of images in deep convolutional neural networks such as VGG19 \cite{simonyan2014very}.
Using information from the activations of the input image and the style image in the neural network, the transfer algorithm can reconstruct an artistic image that shares the same content in the input but with the texture of the style image.  


While the current style transfer model may suffice in creating remixed works of art, 
the boundaries of objects in the resulting image are no longer well defined, and the 
and the message in the original content can be distorted.  For example, if it is applied to a GUI, we may no longer recognize the input and output elements.
Previous research to regain object definition applies imaging processing techniques to the content image such as edge detection~\cite{Li2017Laplacian} or segmentation~\cite{luan2017deep}, and has seen limited success.
Figure~\ref{fig:banner} shows the result of applying the techniques of 
Gatys~\cite{1508.06576}, WCT~\cite{li2017universal}, AdaIn~\cite{huang2017arbitrary}, Lapstyle~\cite{Li2017Laplacian}, Linear~\cite{li2018learning}, and ours to a simple Pizza Hut app and a TikTok photo.  None of the previous techniques retains the boundaries of the buttons in the Pizza Hut app, let alone the labels on the buttons, rendering the display unsuitable as a GUI.  In the street scene, the woman's face is severely distorted, and none of the captions can be read.  

In the original style transfer, the color/texture is captured by the uncentered covariance of features {\em in the same level} in 
the VGG19 model~\cite{simonyan2014very}; to copy the style, the algorithm minimizes the squared error of the covariance between the style and the output images.  The covariance of features represents the joint distribution of pairs of features, and it is not positional.  
We conjecture that since each level is optimized independently, the texture used for the same image position at different levels is chosen independently of each other, thus the lack of positional linkage renders the output incoherent.   This leads us to think that the original style transfer does not fully model style.  

We propose adding a new term to the modeling of style, the {\em uncentered cross-covariance of features across layers}, to link up the style elements chosen for each position across levels.  And we propose adding to style transfer the goal of minimizing the squared error of this term between the {\em style} and output images. Note that prior work to improve object definition focused on copying more information from the {\em content} image.   We show empirically that doing so gives objects well-defined boundaries and better color coherence.  Without doing anything special for text, text and any other detailed objects like logos become legible with this technique. We refer to the cross-covariance as {\em structure}. 
Figure~\ref{fig:banner} shows that our algorithm successfully copies the style, while retaining the details, as all the buttons in the Pizza Hut app are clearly visible and the labels are legible.
Similarly, the details of the woman's face and the caption in the TikTok photo are preserved.  



The contributions of this paper include: 
\begin{enumerate}
\item
ImagineNet: a new style transfer model that can copy the style of art onto images while retaining the details. We add to the traditional modeling of style a new {\em structure} component, which is computed as the {\em uncentered cross-covariance of features across layers} in a VGG neural network. By minimizing the squared loss of the structure component between the style image and the output image, the output has well defined object boundaries and coherent color schemes, unlike the result of the original style transfer.  

\item
The improvements in ImagineNet make possible an important application for style transfer: to improve the aesthetics of applications by restyling them with artwork. ImagineNet is the first style transfer algorithm that can be applied to GUIs without rendering it unusable. We demonstrate three ways to use style transfer on GUIs (a) change the entire look-and-feel of an app by restyling its graphical assets; (b) restyle user-supplied content in an app, (c) restyle an app with dynamically generated content.  

\end{enumerate}

\section{Related Work}
Style transfer is the computer vision task of extracting the style from a reference image and applying it to an input image. Classic style transfer techniques have approached this problem by altering the input’s image statistics to match that of the reference \cite{reinhard2001color} \cite{shih2014style}. Another type of approach, called image analogies, is to identify the relationship between two reference images and applying the relationship on a separate input image to synthesize a new image \cite{hertzmann2001image} \cite{shih2013data}.

With the rising popularity of deep learning, Gatys et al. showed that the internal representations learned by a convolutional neural network (CNN) for image classification tasks can be used to encode the content and style characteristics of an image \cite{1508.06576}. This influential work significantly advanced the field of style transfer in two ways. First, it showed that the texture and color style of an image can be encapsulated by computing the uncentered covariance of an image’s feature maps extracted from a CNN. The feature map representations have even been used with the classic image analogy style transfer approach \cite{liao2017visual}. Second, it introduced an iterative optimization algorithm, called neural style transfer, to synthesize images that contain the content of one reference image, and the style of another. 

Both fronts have then been expanded upon to further improve neural style transfer type of algorithms. On the style representation front, Li and Wand showed that Markov Random Fields can be used as an alternative to the covariance matrix to capture style \cite{li2016combining}. Furthermore, Li et al. showed that generating an image to match the covariance matrix of the style image’s feature maps is actually equivalent to matching the statistical distribution of the style image’s feature maps as computed by the maximum mean discrepancy function \cite{li2017demystifying}. This work lead to an alternate view of neural style transfer as a statistical feature transfer problem.

On the algorithmic front, several works expanded upon the original optimization-based neural style transfer. A drawback of the optimization approach is its speed, which cannot run in real-time. Johnson et al. trained an image transformation neural network that can perform style transfer with a single forward pass of the input image through the network, allowing real-time stylization but at the cost that the network can only have one fixed style \cite{1603.08155}. Subsequent works have created real-time arbitrary style transfer neural networks by using the statistical feature transfer formulation \cite{huang2017arbitrary} \cite{chen2016fast} \cite{li2017universal} \cite{li2018learning}. While these networks are optimized for speed and without the limitation of a fixed style, improving the visual quality of the generated images is still an active area of research.

There have been numerous methods to improve on each of the aforementioned neural style transfer algorithms. Spatial information and semantic masks have been used to constrain transfer of style in corresponding regions \cite{champandard2016semantic} \cite{gatys2017controlling}. For photorealistic applications of style transfer, several methods use local image statistics and scene information to prevent distortions in the generated image \cite{luan2017deep} \cite{mechrez2017photorealistic} \cite{yang2019multi} \cite{li2018closed}. Other works attempt to preserve lines and edges in the generated image guided by Laplacian operators \cite{Li2017Laplacian} and depth maps of the reference and generated images \cite{liu2017depth}.  
ImagineNet is the only proposal that augments the original neural formulation with a new loss term, with the goal to improve the definition of the output image.  Furthermore, unlike other techniques, ours copy more information from the style image rather than the content image.

\section{The Style Transfer Model}

\begin{figure}
\centering
  \includegraphics[width=0.8\columnwidth]{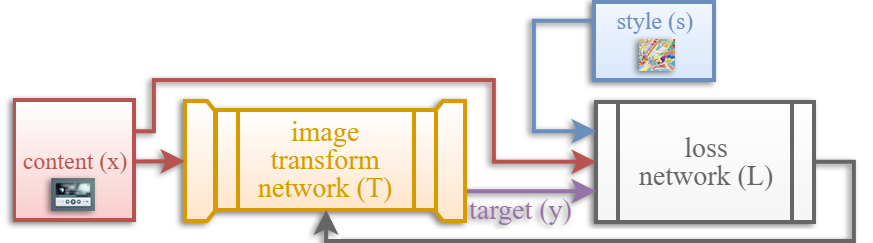}
  \caption{
  	Diagram of the training process. At test time, only the image tranform network is used. 
  }~\label{fig:figure_pipeline}
\vspace{-.3in}
\end{figure}

Our style transfer algorithm is based on the feed-forward image transformation network~\cite{1603.08155}, as shown in Figure~\ref{fig:figure_pipeline}. Note that our system consists of two neural networks, the image transformation network that takes an input image and applies a particular style to it, and a fixed loss network used to extract features between the input image and generated image for a loss function we describe later.
The fixed loss network we use is VGG19~\cite{SimonyanZ14a} pre-trained on the ImageNet dataset~\cite{russakovsky2015imagenet}.

The image transformation network, $T$ is a convolutional encoder-decoder network with residual connections~\cite{radford2015unsupervised,odena2016deconvolution}. For a specific style reference image $s$ and an input content reference image $x$, the network produces an output image $y$ in a single forward-pass. We train each transformation network with one pre-selected reference style image, we use images from the entire COCO dataset\cite{lin2014microsoft} as reference content images.
The network minimizes the following loss function with respect to the input reference images $s,x$ and the generated image $y$: 

\begin{align*}
 L =  & \lambda_1 \Lcontent(x,y) + \lambda_2 \Ltexture(s,y) + \\
& \lambda_3 \Lstructure(s,y) + \lambda_4 \LTV(y)
\end{align*}

The formula above is identical to the traditional formulation, except for the introduction of $\Lstructure$, the structure loss term, which we will discuss last.  $\Lcontent$ copies the content from the input image, $\Ltexture$ copies the color and texture of the style image, and $\LTV$ is the  total variation regularization \cite{mahendran2015understanding} to the loss function to ensure smoothness in the generated image~\cite{1508.06576,1603.08155}.

\subsection{Content Component}
The feature maps in layer $\ell$ of VGG19 of image $x$ are denoted by $\mathbf{F}^\ell(x) \in \mathbf{R}^{C_l \times N_l}$, where $C_l$ is the number of channels in layer $\ell$ and $N_\ell$ is the height times the width of the feature map in layer $\ell$. 
$F_{i,j}^\ell(x)$ is the activation of the $i$-th channel at position $j$ in layer $\ell$ for input image $x$, 
To ensure that the generated image $y$ retains the content of the input image $x$, the content loss function $\Lcontent(x,y)$ is used to minimize the square error between the features maps extracted from the high-level layers in the VGG19 loss network for $x$ and $y$.
That is, 
$$\Lcontent(x, y) = \sum_{i=1}^{C_\ell} \sum_{j=1}^{N_\ell} (F^\ell_{i,j}(x) - F^\ell_{i,j}(y))^2$$ 

\subsection {Texture Component}

Gatys et al. propose that the colors and textures of an image can be captured by the uncentered covariance of features in each layer: $\mathbf{G}^\ell(x) \in \mathbf{R}^{C_\ell \times C_\ell}$, 
where $C_\ell$ is the number of channels in layer $\ell$:  
$$G_{i,j}^{\ell}(x) = \displaystyle \sum_{k=1}^{N_\ell} F^\ell_{i,k}(x)F^{\ell}_{j,k}(x)$$
$G_{i,j}^{\ell}$ is the joint probability of feature $i$ colocating with feature $j$ in layer $\ell$. 
Gatys et al. show layers with similar texture and colors have similar covariance. 
To copy the style, Gatys proposed to minimize the squared error between the covariance of the style and output images in the low levels:  
$$\Ltexture(s, y) = \displaystyle\sum_{\ell\in \text{layers}} \beta_{\ell}\sum_{i=1}^{C_\ell} \sum_{j=1}^{C_\ell} (G^\ell_{i,j}(s) - G^\ell_{i,j}(y))^2$$ 

As the number of channels increases in higher levels, so do 
the dimensions of the covariance.  The dimensions of the covariance matrix are $64 \times 64$ for layers 1 and 2, $128 \times 128$ for layers 3 and 4, and 
$512 \times 512$ for layers 4, 5, 6, and 7.





We see from Figure 1 that the style image has well-defined shapes and boundaries, yet the original style transfers generate images that lack definition. The object boundaries are no longer sharp and the colors and textures are no longer coherent within an object.  

Texture and color are captured as covariance of features in the same level, and this information is not positional.  It is possible to generate many output images that have equivalent differences in covariance with the style image.  The choices made in each layer to minimize style loss are independent of each other.  Content loss provides the only cross-layer constraints in the original formulation.
The content representation at the high layer of the convolutional neural network, however, is coarse, since the convolution filters at the upper layer cover a large spatial area.
The image transformer has the freedom to choose textures as long as the objects are still recognizable at coarse granularity.
Boundaries are lost in this representation because the same object can be recognized whether the edge is straight or blurry or jagged, or if the line is drawn with different colors. 

\subsection{Structure Component}

We introduce the concept of {\em uncentered cross-covariance of features across layers}, referred to as {\em structure}, to model the correlation of texture across layers.  Our objective is to further constrain the output image so it has similar cross-covariance as the {\em style} image.



Lower-level layers have less channels but more data points than higher-level layers.  To compute the cross-covariance, we give the feature maps of different levels the same shape by upsampling the lower-level feature maps along the spatial dimension, and upsampling the higher-level feature maps in the channel dimension, by duplicating values.  The upsampled feature map of layer $\ell_1$ to match layer $\ell_2$ for image $x$ is denoted by $\mathbf{F}^{\ell_1,\ell_2}(x) \in \mathbf{R}^{\text{max}(C_{\ell_1}, C_{\ell_2}) \times \text{max}(N_{\ell_1}, N_{\ell_2})}$.

The uncentered cross-covariance of feature maps between layers $\ell_1$ and $\ell_2$, $\ell_1 < \ell_2$, is denoted by 
$\mathbf{G}^{\ell_1,\ell_2}(x) \in \mathbf{R}^{C_{\ell_2} \times C_{\ell_2}}$, where $C_{l_2}$ is the number of channels in layer $\ell_2$. The cross-covariance computes the joint probability of a pair of features belonging to different layers.

$$G^{\ell_1,\ell_2}_{i,j}(x) = \displaystyle \sum_{k=1}^{N_{\ell_1}} F^{\ell_1,\ell_2}_{i,k}(x)F^{\ell_2,\ell_1}_{j,k}(x)$$

To copy the covariance to the output image, we introduce a {\em structure loss} function to minimize the squared error between the cross-covariance of the style and output images:
$$\Lstructure(s, y) = \displaystyle\sum_{\ell_1,\ell_2\in \text{layers}} \gamma_{\ell_1,\ell_2} \sum_{i=1}^{C_{\ell_2}} \sum_{j=1}^{C_{\ell_2}} (G^{\ell_1,\ell_2}_{i,j}(s) - G^{\ell_1,\ell_2}_{i,j}(y))^2$$

\noindent where $\ell_1 < \ell_2$ and  $\gamma_{\ell_1,\ell_2}$ is the loss weight factor.
The structure loss term constrains the choice of textures across layers in the output image to match the correlation between layers in the style image, thus copying the style more faithfully. 


\begin{figure}[htb]
\centering
  \includegraphics[width=\columnwidth]{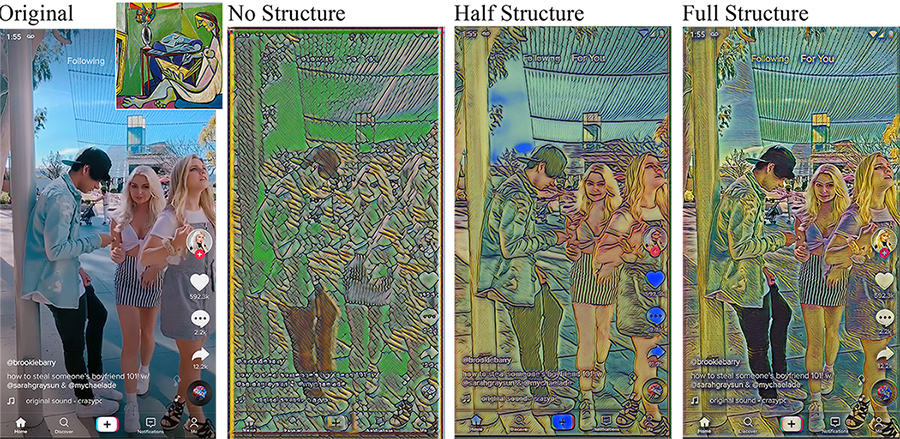}
  \caption{
An ablation example on structure loss: 
a Tiktok picture restyled with no structure, half structure, and full structure. 
The structure and legibility improve as the structure loss term increases.}~\label{fig:without_structure}
\end{figure}

As an example, Figure~\ref{fig:without_structure} shows the effect of adding the cross-covariance loss term to style transfer by applying La Muse to a TikTok photo.  Without cross-covariance, the result is similar to that from Johnson\cite{1603.08155}.  It is hard to pick out the women in the photo. With cross-covariance, the people and even the net overhead are visible easily; it has interesting colors and textures; the words in the image are legible.  When we half the weight of the cross-covariance error in the loss function, we see an intermediate result where the structure starts to emerge, hence we coined this component ``structure''.

 


\subsection{Ensuring Target Image Clarity}
\short{
Fine grain lines and text is an important element across many images.
However, traditional style transfer does not preserve the clarity of these assets in the content image.
For example, text is treated and stylized the same as other visual elements and the results are hard-to-read.
}

To understand how edges are learned and represented in a CNN,
we visualize the neuron activation maps of VGG19 to identify the layers of neurons salient to text in an image.
Neuron activation maps are plots of the activations during the forward propagation of an image.
We sample random images from ImageNet classes and overlay objects on it.
We then forward pass this combination image into VGG19 to obtain the activations, and identify the layers with the highest activations with respect to outlines of objects.

VGG has 5 convolution groups, each of which consists of 2 to 4 convolution-and-relu pairs, followed by a max pooling layer. A relu layer is numbered with two indices, the first refers to the group, and the second the number within the group; so, relu3\_1 refers to the first relu in group 3.  

From our visualizations and analysis, we observe that neurons in the ``relu1\_1'' and ``relu3\_1''  layers of VGG carry the most salient representations of image outlines.
Earlier layers in image classification neural networks tend to be edge and blob detectors, while later layers are texture and part detectors \cite{Lee:2009:CDB:1553374.1553453}.
These results suggest that sufficient weight must be placed in these layers so that the outline information is carried to the output image.

\begin{figure}[htb]
\centering
  \includegraphics[width=0.8\columnwidth]{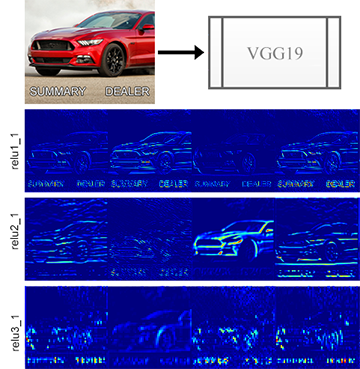}
  \caption{ 
Layers ``relu1\_1'' and ``relu3\_1'' in the VGG19 network are highly responsive to text within images. 
  }~\label{fig:figure_activations}
\vspace{-.2in}
\end{figure}

\begin{table}[htb]
\small
  \centering
  \begin{tabular}{l | l | r}
    \hline
    \hline
    Loss Function & Layer & Weight \\
    \hline
$\Lcontent$ &   ``relu4\_1'' & 5.6 \\
\hline
$\Ltexture$ & ``relu1\_1''    & 1.1 \\
 &  ``relu2\_1''    & 1.3 \\
 &  ``relu3\_1''    & .5 \\
 &  ``relu4\_1''    & 1.0 \\
\hline
$\Lstructure$ & ``relu1\_1'' $\times$ ``relu1\_2''    & 1.5 \\
 &  ``relu1\_1'' $\times$ ``relu2\_1''    & 1.5 \\
 &  ``relu2\_1'' $\times$ ``relu2\_2''    & 1.5 \\
 &  ``relu2\_1'' $\times$ ``relu3\_1''    & 1.5 \\
\hline
    $\LTV$  &    & 150 \\
\hline
  \end{tabular}
  \caption{Weights for the loss functions.}~\label{tab:hyperparameters}
\vspace{-.3in}
\end{table}

\begin{table*}[htb]
  \includegraphics[width=2\columnwidth]{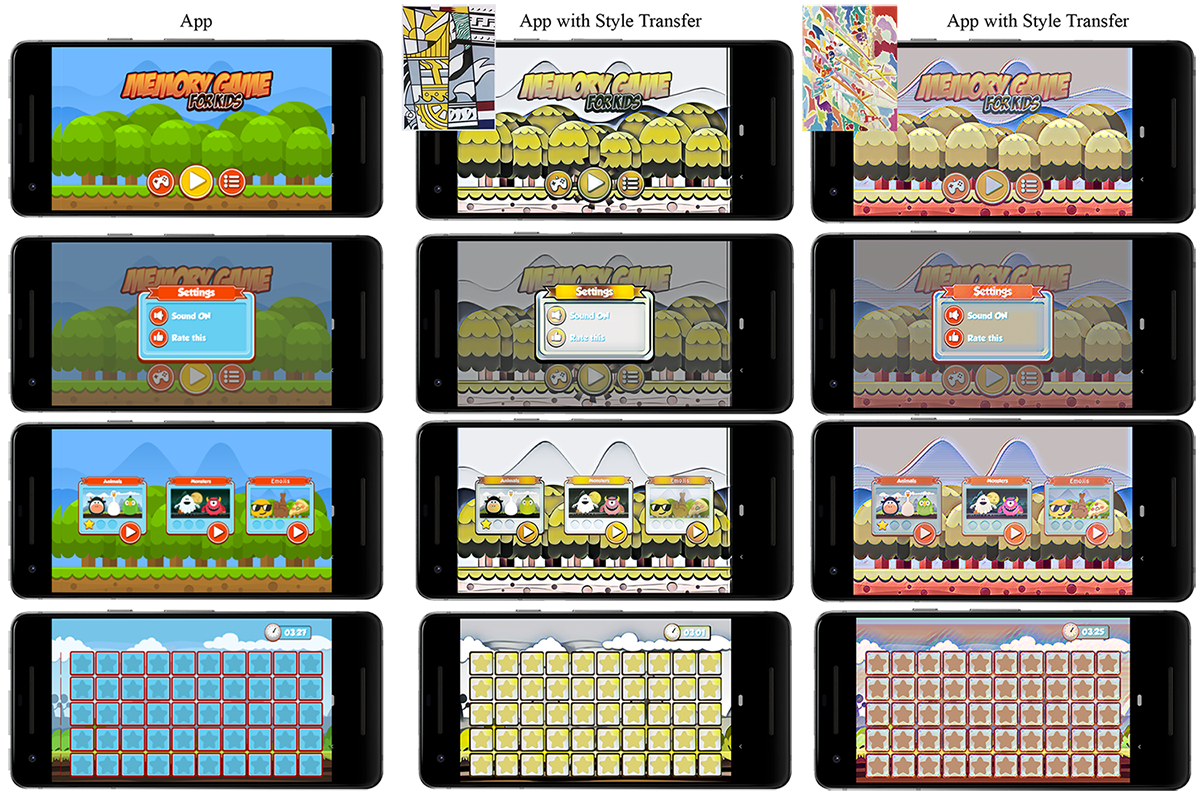}
\caption{
Example of how ImagineNet can be used to style assets in the Memory Game app to create 
a fully functional app. 
}~\label{table:memory}

\end{table*}

\subsection{Functionality and Aesthetics Tradeoff}


We tuned the hyperparameters of the loss functions through a binary search.  
By evaluating the text legibility, edge distinctions, and distortions in the output obtained from 20 diverse pairs of content and style images, we derived the parameters shown in Table \ref{tab:hyperparameters}.  
As shown in the table, the content loss term is calculated on the feature maps in group 4.  

When too much weight is placed on the lower VGG convolution layers in the loss function, only color and smaller elements of styles are transferred over, and the results look flat.
Due to this, we increase loss weights in convolutional groups 3 and 4.
However, we observe that lines begin to blur in the results.
Finally, we increase the weight of the structural loss, and arrive at a model that transfers global style and maintains text clarity.
To determine the total variational loss weight, we iteratively decrease the weight until deconvolutional artifacts--such as the checkerboarding effect--begin to appear.  



\section{How to Restyle Apps}

In this section, we show how style transfer can be applied to three kinds of mobile apps: apps that use a static set of graphical assets, apps that accept user graphical contributions, and finally apps that  generate interfaces dynamically. 

\subsection{Graphical Asset Restyling}

The look-and-feel of many applications is defined by a set of graphical assets stored in separate files.  For such apps, we can generate a new visual theme for the app by simply replacing the set of assets with those restyled according to a given style.  

Apps for mobile devices typically separate out their graphics into a separate folder called ``assets''.
The assets folder contains the sets of similar images for different resolutions, themes, or geographic localities.  The text, such as the current time or the number of points a user scored, is dynamically generated and overlaid on top of the images.  
Thus, we can create a fully functional app, with a different style, by simply replacing the files in the ``asset'' folder with restyled images. 
Since assets are dynamically loaded as the app runs, this process does not require any functionality changes in the app source code.

To restyle an app, we first apply ImagineNet to the assets for a chosen style image.
Next, we clean up the border in a post-processing step.
Unless the graphics in the asset is rectangular, it carries a transparency mask to indicate the border of the asset.   We copy over the same transparency mask to the restyled image.  This is necessary because the convolution-based neural network have padding and stride parameters that generate additional pixels outside the original pixels. 


To demonstrate this process, we use an open-source Android app called Memory Game~\cite{memorygame}. 
We unpack the apk, extract the assets, apply style transfer to all assets, and then resign the app.  Various screen shots of this style-transferred app are shown in Table~\ref{table:memory}. Here we show the results of two automatically generated, fully functional versions of Memory Game with different look-and-feel.
ImagineNet reproduces the styles of the artwork and yet retains the full details of the images, making the game perfectly playable. 

\begin{figure}[htb]
\centering
\includegraphics[height=10cm]{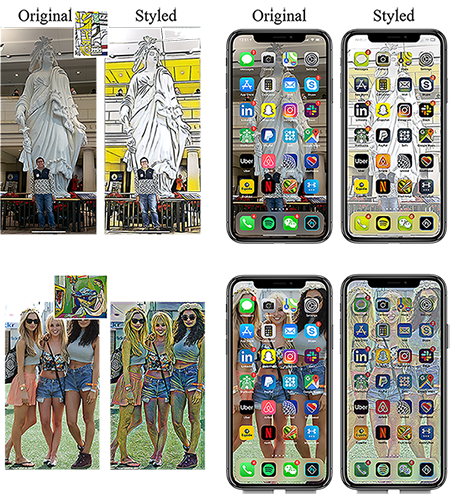}
\caption{Style transfer applied to the iOS home screen.  We show, from left to right, a given style image, a user-supplied photo, a re-styled photo given the style, the original homescreen, and the new theme with re-styled icons and the re-styled photo as a background. Best viewed digitally in full resolution.}
\label{fig:ui}
\end{figure}

\subsection{User-Supplied Content}

Consumers have shown a lot of interest in apps that let them personalize their content; examples include applying filters in Instagram to enhance their photos and using artistic style transfers in Prisma and DeepArt.  As demonstrated in Figure~\ref{fig:without_structure}, our style transfer can faithfully transfer the style of the art while preserving the message of the original message.  Here we illustrate how we can take advantage of this property to provide personalization, using launchers as an example. 

A launcher is a mobile app that changes the app icons and the background to a different theme.  Launchers are popular among mobile phone users; over a million different themes are available, many for purchase. 
Style transfer offers an automatic technique to generate new themes based on masterpiece artwork.  
As an automatic procedure, it can also support users' provided content.  
In particular, mobile users often want to use personal photos as the background of their phones, we restyle their background in the same artistic style as the icon theme to create a stylishly consistent home screen experience.

The algorithm consists of two parts.  The first is to statically restyle the icons, using the same algorithm as described above.  The second is to restyle a user-supplied image for the background.  The time and cost in restyling an image are negligible as discussed in Section~\ref{sec:speed}.  The installer can then accept these assets and draw the homescreen accordingly.  

We show two examples of this process in 
Figure \ref{fig:ui}.  We show the results of applying a given style to the background, then the result of icons on the background.  The results show that the re-styled backgrounds retain the message in the original images.  Moreover, the restyle icons match the backgrounds.  We reduce opacity on the background pictures to make the icons and names of the apps more visible. 



\subsection{Dynamically Restyling an App}

Some apps generate graphical displays dynamically. For example, a map app needs to present different views according to the user input on the location as well as the zoom factor.  Such dynamic content needs to be dynamically re-styled as they are generated.  Here we use 
Brassau, a graphical virtual assistant~\cite{Fischer2018Brassaua}, as an extreme case study, since it generates its GUI dynamically and automatically. 

\begin{figure}
\centering
  \includegraphics[width=0.9\columnwidth]{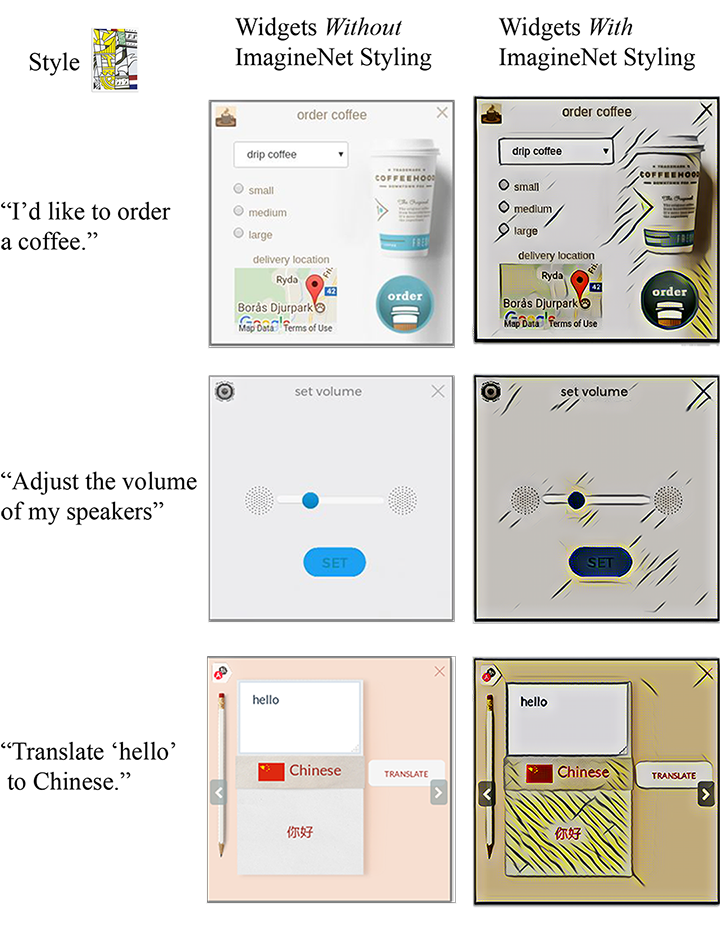}
  \caption{ImagineNet applied to Brassau to generates widgets from natural language commands (column 1) and restyles them, in this case using Lichtenstein's Bicentennial Print (column 2).
  }~\label{fig:brassau}
\vspace{-.3in}
\end{figure}

Brassau accepts user natural language input, drawn from an extensible library of virtual assistant commands, and automatically generates a widget.  For example, the command such as ``adjust the volume of my speakers'' would generate a widget that lets users control the volume with a slider. 
Brassau has a database of GUI templates that it matches to commands users issue to a virtual assistant.
However, the templates in the database come from a variety of designers and the styles across templates are inconsistent. As a result, the widgets generated, 
as shown in Figure~\ref{fig:brassau}, are inconsistent in style.  A developer of Brassau, not an author of this paper, uses ImagineNet to give its dynamically widgets a uniform style according to its user's preference. 

To support interactivity, the developer separates Brassau's GUI into three layers.
The bottom layer contains the original display, including the buttons and background templates.
The middle layer, hiding the bottom layer, is the stylized image, obtained by rasterizing the bottom layer with RasterizeHTML and stylizing it via the ImagineNet server.  
The top layer has user inputs and outputs, 
providing real-time GUI interaction.
Button clicking is supported by the bottom layer, as the top two layers are made transparent to clicking.
It took the developer 8 hours to add ImagineNet to Brassau, by specifying the  ``absolute positioning'' and ``z-index'' CSS properties of the user interface elements.

In a pilot study with 8 participants, 4 male and 4 female, 6 
preferred the stylized application over the original, 1 was indifferent, and 1 liked it less.
Users who liked the styled app said that they liked the consistency in style across apps, with one commenting ``now it looks like it was all designed by one person instead of a group''.
They all cited legibility as the highest priority in their choice of styles. 


\begin{table*}
\centering
\includegraphics[width=1.8\columnwidth]{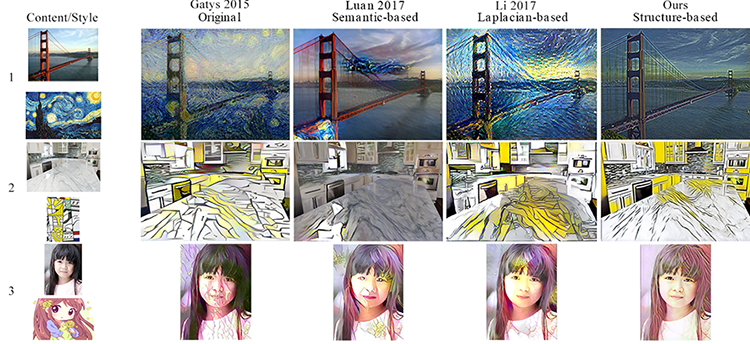}
\caption{Photorealistic style transfer using, from left to right, the original style transfer algorithm, semantic-based style transfer, Laplacian-based style transfer, and ours.
}
\label{table:photorealistic}
\vspace{-.2in}
\end{table*}

\section{Comparison of Style Transfer Techniques}

In this section, we discuss how ImagineNet compares with previous style transfer algorithms on style transfer for photorealistic images and for GUIs. 

\subsection{Photorealistic Style Transfer}
From our literature review, we did not find any other works attempting style transfer on GUIs. The closest prior work is style transfer on photo-realistic images, where the generated images are expected to look like a natural, realistic image. This work is the most relevant to ours because of the expectation to maintain lines and edges without distortions in the final image.

Table ~\ref{table:photorealistic} shows the results of the original style transfer algorithm~\cite{1508.06576}, semantic-based style transfer~\cite{1508.06576}, Laplacian-based style transfer~\cite{Li2017Laplacian}, and ours on three pairs of photos and style images. We chose these works for comparison since the authors specifically target the photo-realistic domain.

The first is a typical example used in photo-realistic style transfer where the style of Van Gogh's Starry Night is copied to a photo of the Golden Gate bridge. ImagineNet produces the only output image that shows the bridge cables clearly, demonstrating how a purely neural algorithm is more robust than using the Laplacian for edge detection. Additionally, ImagineNet is not prone to segmentation failures, which is observed in the semantic-based result.  However, ImagineNet copies mainly the color scheme, as Van Gogh's signature style is in conflict with the priority of preserving details. 

In the second example, we choose
a real-world image with reflections as a challenging content image and pick for style the 
Lichtenstein's Bicentennial Print which has distinct lines, as a contrast to Van Gogh's Starry night. 
A good style transfer will copy the clean lines from the style to this photo. 
It can be observed that the original and Laplacian-based algorithms fail to contain style within the boundaries of the countertop. While the semantic-based model is able to maintain the boundaries, it only transfers a small amount of style to a corner of the image, which is  due to a failed segmentation of the content image. Ours, on the other, has clean straight lines for the kitchen table and counters, and also copies the style to the table top well.  

In the third example, we use source images from Li et al.~\cite{Li2017Laplacian} for comparison. ImagineNet is the only technique that does not disfigure the girl's face or hair, while faithfully changing her hair to match that of the style image. This example illustrates that our algorithm adapts to the structure of images and does not produce overly-styled images when the inputs are simple.

\subsection{Comparison with Style Transfers on GUIs}

We applied former style transfer techniques to all the input pairs shown in Fig.~\ref{table:style}, and none succeeded in preserving the details of a GUI.  Due to space constraint, we can only show two representative results in Figure~\ref{fig:banner}.  
When we apply ``La danse, Bacchante'' to the Pizza Hut app, we see that ImagineNet is the only technique that makes the logo ``Pizza Hut'' recognizable; it transfers the mosaic texture properly to the background; the buttons are well defined; and the letters on the buttons are legible. 
When we apply ``La Muse'' to a Tiktok picture, the woman's face is distorted by 
all the other style transfers; none of the writing can be read;  none of the people in the background are recognizable.  ImagineNet carries all the details, while adding texture whenever possible without ruining the message, as shown by the lines on the ground and the man's pants.

\begin{figure}
\centering
  \includegraphics[width=0.9\columnwidth]{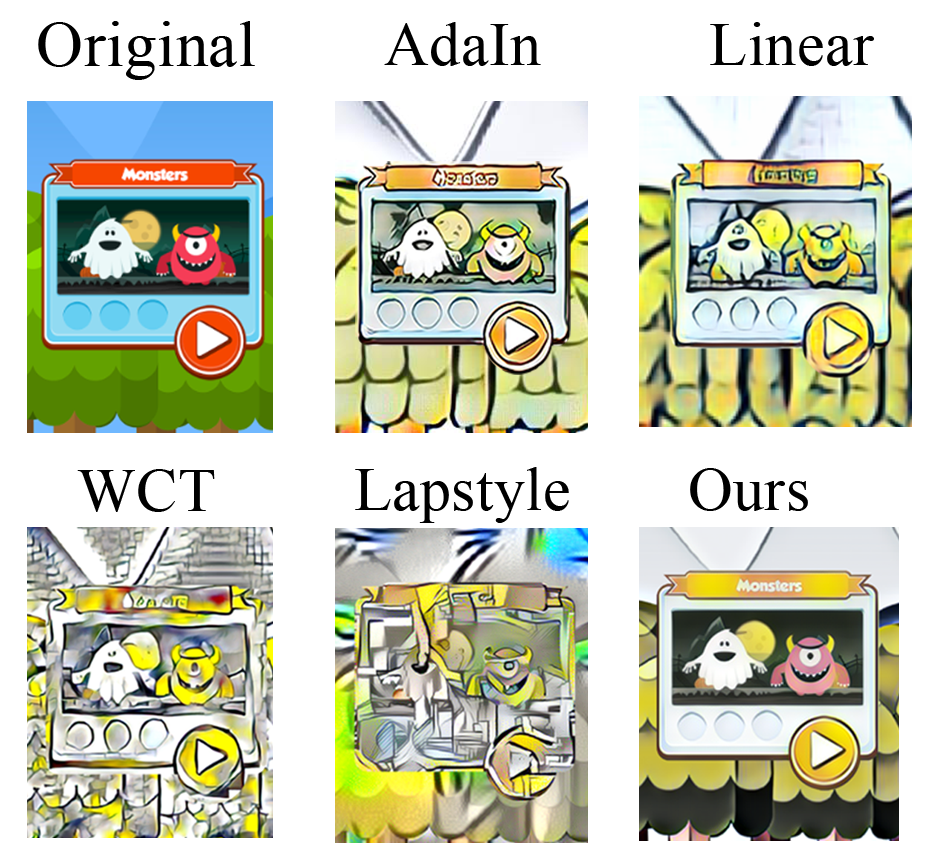}
  \caption{Applying different style transfers to the Memory Game; only ImagineNet can make the words legible.}~\label{fig:comparisons_asset}
\vspace{-.3in}
\end{figure}

We also applied the former techniques to the MemoryGame to understand if they can be used for coloring assets.  We use the Bicentennial Print as the input style as its clean style makes it easiest to transfer. 
Fig.~\ref{fig:comparisons_asset} shows that none of the games styled by other techniques are playable.  The best result is obtained by Linear~\cite{li2018learning}, but the words are not legible.   

\begin{table}[]
\centering
\small
\label{results_description_style}
\begin{tabular}{p{3cm}| p{5cm} }
\hline
\hline
\textbf{Style}                                                                  & \textbf{Style Description}                       \\
\hline
\begin{tabular}[t]{@{}l@{}}\textit{La Muse}~\cite{Amuse19383:online} \\ Pablo Picasso 1935\end{tabular}               & Cubism, Surrealism. Bold, colorful, whimsical.     \\ \hline
\begin{tabular}[t]{@{}l@{}}\textit{La danse, Bacchante}~\cite{Bacchant46:online} \\ Jean Metzinger, 1906\end{tabular}  & Divisionism. Distinct mosaic pattern, pastel colors with blue borders around the main figure.            \\ \hline
\begin{tabular}[t]{@{}l@{}}\textit{Les r\'egates}\cite{Lesregat17:online}\\ Charles Lapicque, 1898\end{tabular}                          & Fauvism, Tachisme. Colorful with small contiguous areas with white borders                                            \\ \hline
\begin{tabular}[t]{@{}l@{}}\textit{Spiral Composition}~\cite{SpiralCo6:online}\\ Alexander Calder, 1970\end{tabular}                           & Abstract. A strong yellow area, complemented with blue, red, and white concentric circles with black radials. \\ \hline
\begin{tabular}[t]{@{}l@{}}\textit{Bicentennial Print}~\cite{Bicenten51:online}\\ Roy Lichtenstein, 1975\end{tabular}   & Abstract Expressionism. Strong geometric shapes with yellow highlights.                                                    \\ \hline
\begin{tabular}[t]{@{}l@{}}\textit{A Lover}~\cite{ALoverTo78:online}\\ Toyen, unknown \end{tabular}                           & Expressionism. Simple design and nearly mono-chromatic.                        \\
\hline
\hline
\end{tabular}
\caption{Description of styles used in Table~\ref{table:megatable}.}~\label{table:style}

\end{table}

\begin{table*}
  \includegraphics[width=2\columnwidth]{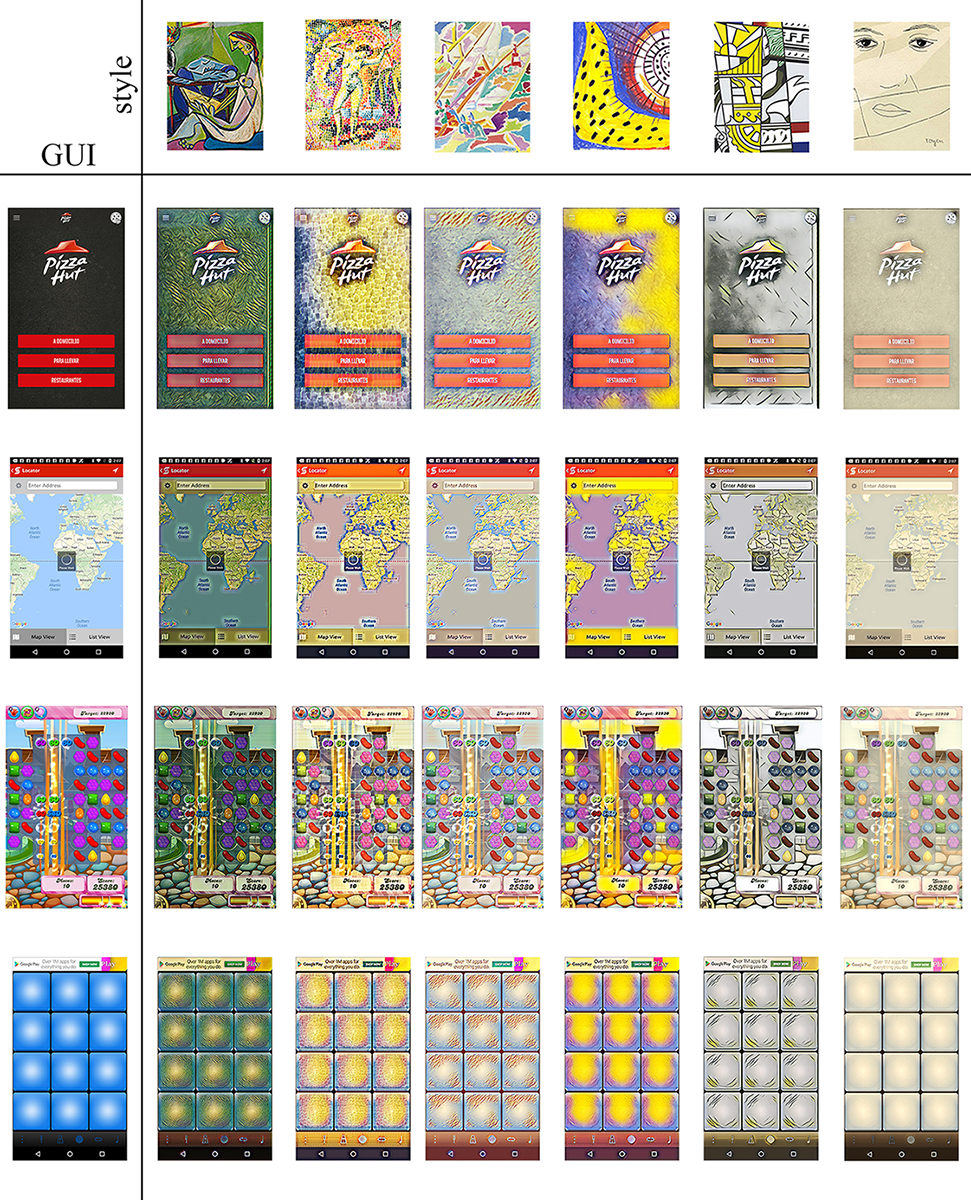}
\caption{Results of transferring the style of the paintings from Table~\ref{table:style} to GUIs. The details can best be viewed on a computer.
}~\label{table:megatable}

\end{table*}

\section{Experimentation with ImagineNet}

\subsection{Results of ImagineNet on Apps}

Here we show how the results of applying ImagineNet to a wide range of apps. For these, as we do not have access to the sources, we simply show the effect of applying ImagineNet to the images captured on the mobile devices.  
In Table \ref{table:megatable}, the select apps are shown on the left column; the styles are shown on the top.
ImagineNet shows detailed information clearly, while still showing the style of the original artwork.
ImagineNet handles the high information density across different kinds of GUI elements: text, icons, country boundaries, and gaming assets.
Note how the music keys of the last GUI are all restyled the same way, providing coherence across the whole app. 

\subsection{Quality of Results Study}

We conduct a user study to see the quality of ImagineNet results.
We present GUIs generated by style transfer algorithms from Gatys et al., Johnson et al., Luan et al., Li et al., and ones generated by ImagineNet to 31 female and 19 male Mechanical Turk evaluators in the U.S.
All evaluators selected the GUIs generated by ImagineNet as their preferred GUI.
To assess  if ImagineNet maintains text legibility, we present evaluators with 6 styled GUIs and ask them to transcribe all text in the GUI.
All evaluators retyped the text with 100\% accuracy.
In addition, to show how dutifully the style is maintained in the new image, we show evaluators 6 style images and 6 stylized GUIs, and have them match the two sets.
Evaluators match the style to the transferred GUI with 79\% accuracy.

\begin{figure}
\centering
  \includegraphics[width=0.9\columnwidth]{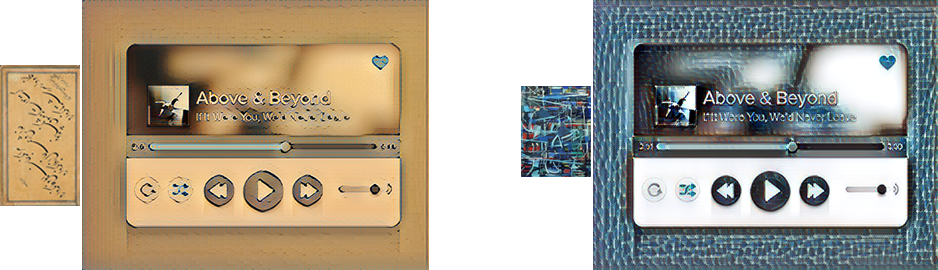}
  \caption{
  	Examples of illegible GUIs generated by ImagineNet.
  	The style images, shown below the GUIs, lack structural elements in the style to make letters legible.
  }~\label{fig:errors}
\end{figure}

\subsection{Style Image Selection Study}
To understand if artwork can be used on apps in general, we conduct a user study on the results of applying ImagineNet on a random set of artwork and app images.
We train ImagineNet for 96 randomly selected images, under different style classifications in the Paint by Numbers dataset~\cite{paintbynumbers}. 
We apply these styles to a randomly sampled images to create 4800 styled images.
We show to each of the 48 Mechanical Turk evaluators 180 randomly sampled styled GUIs, and asked them to rate the design on a Likert scale from 1 (bad) to 5 (good). 
The average rating of the 96 styles ranges from 1.6 to 4.5, with only 14 rated below 3; the remaining 82 paintings are distributed uniformly between 3 and 4.5.   The standard deviation of ratings received for each style ranges from 0.5 to 1.4, with an average of 1.1, suggesting that there is a general agreement whether a style is good and it depends less on the content to which it is applied.

We found that styles classified as \textit{Abstract Expressionism}, \textit{High Renaissance}, \textit{Modernismo}, and \textit{Pop Art} transfer well due to clear edges and vibrant colors generally present in these paintings.
On the other hand, \textit{Action painting}, \textit{Lettrism}, \textit{Kinetic Art}, and \textit{Purism}, did not transfer well.  

Not all styles are suitable for restyling apps. 
Since ImagineNet transfers the structure component in style, if the style image has no clear object borders anywhere, it will not be able to provide a crisp output.
Styles with faded colors, with less than three colors, or containing little texture were found to not work well.  
Figure~\ref{fig:errors} shows representative styles that do not transfer well to GUIs.
The style transfers well but the text is illegible because the style image lacks the structural elements needed for lettering.
This result suggests that our structure component should indeed be included as a component of style.

\begin{figure}
\centering
  \includegraphics[width=0.9\columnwidth]{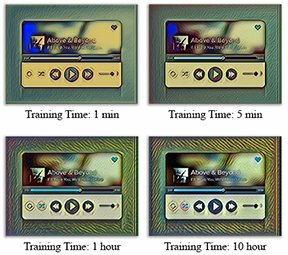}
  \caption{
  	While training, a user can cancel training within one minute if the generated result does not look promising. 
  	After 5 minutes, 1 hour, and 10 hours the style becomes clear.
  }~\label{fig:training_time}
\end{figure}

\subsection{Speed of ImagineNet}
\label{sec:speed}
All the speeds reported are measured on a computer with 6 core 3.50 GHz processor with 48 GB of RAM and the following GPUs: Nvidia Titan V 12GB, Nvidia Titan X 12GB and Nvidia GTX 980Ti 6GB. 
The code is written in Python 3.5 and TensorFlow 1.5 using CUDA V9.0 and cuDNN V6.0.

For the best qualitative results, we train using batch size of 1, for 10 hours, but a preliminary approximation can be obtained in 1 minute, shown in Figure~\ref{fig:training_time}.
Once the image transformer is trained for a style, the image transformation time only depends on the size of the input image.
We test the run-time for different sizes of color GUIs.
On average over five runs, a 512x288 sized image takes 3.02 ms while a 1920x1080 sized image takes 104.02 ms. 
On average, the transformation took 0.026 $\mu$s/pixel.

\section{Limitations}
By prioritizing legibility, ImagineNet cannot generate images as artistic as those produced by prior techniques. It is possible to adjust the weight of the structure loss term to create a different balance between aesthetics and functionality. 

Many algorithms have been proposed recently to speed up style transfer~\cite{li2017universal, li2018learning}.  Since the structure loss term is computed across layers, ImagineNet is not immediately amenable to such optimization techniques. The algorithm as is cannot be run in real-time on mobile phones. More work is required to optimize ImagineNet in the future.  
\section{Conclusion}

We propose ImagineNet, a neural network that transfers the style of an image to a content image, while retaining the details of the content.  While other attempts to retain details in content add processing steps to the {\em content}, our model copies more information from the {\em style}.  We show that the original formulation of style is missing a structure component, computed as the uncentered cross-covariance between features across different layers of a CNN.  By minimizing the squared error in the structure between the style and output images, 
ImagineNet retains structure while generating results with texture in the background, shadow and contrast in the borders, consistency of design across edges, and an overall cohesiveness to the design.

We show how ImagineNet can be used to re-style a variety of apps, from those using a static set of assets, those incorporating user content, and to those generating the graphical user interface on-the-fly.  By style-transferring well-chosen artwork to mobile apps, we can get beautiful results that look very different from what are commercially available today.

\section{Acknowledgments}
This work is supported in part by the National Science Foundation under Grant No. 1900638 and the Stanford MobiSocial Laboratory, sponsored by AVG, Google, HTC, Hitachi, ING Direct, Nokia, Samsung, Sony Ericsson, and UST Global.

\balance{}

\bibliography{sample}

\end{document}